\documentclass[format=acmsmall, review=false, screen=true]{acmart}

\usepackage{booktabs} 
\usepackage{subfig}
\usepackage[ruled]{algorithm2e} 
\usepackage{soul}

\newcommand{\Fscore}{\ensuremath{\textit{F-score}}}%
\graphicspath{{./figures/}}

\SetAlFnt{\small}
\SetAlCapFnt{\small}
\SetAlCapNameFnt{\small}
\SetAlCapHSkip{0pt}
\IncMargin{-\parindent}

\acmJournal{TIST}
\acmVolume{9}
\acmNumber{4}
\acmArticle{39}
\acmYear{2010}
\acmMonth{3}
\copyrightyear{2009}

\setcopyright{acmlicensed}

\acmDOI{0000001.0000001}

\received{February 2007}
\received[revised]{March 2009}
\received[accepted]{June 2009}

\begin{document}
\title[Multi-Objective Variational Autoencoder]{ Multi-Objective Variational Autoencoder: an Application for  Smart Infrastructure Maintenance}  

\author{Ali Anaissi}
\affiliation{%
 \institution{School of Computer Science, The University of Sydney}
  	\city{Sydney}
  	\state{NSW}
  	\postcode{2006}
  	\country{Australia}}
\email{ali.anaissi@sydney.edu.au}

\author{Seid Miad Zandavi}
\affiliation{%
  \institution{School of Computer Science, The University of Sydney}
  \city{Sydney}
   \state{NSW}
  \postcode{2006}
  \country{Australia}
}
\email{miad.zandavi@sydney.edu.au}

\begin{abstract}
	
		Multi-way data analysis has become an essential tool for capturing underlying structures in higher-order data sets where standard two-way 	analysis techniques often fail to discover the hidden correlations between variables in multi-way data. We propose a multi-objective variational autoencoder (MVA) method for smart infrastructure damage  detection and diagnosis in multi-way sensing data  based on the reconstruction probability of  autoencoder deep neural network (ADNN). 	Our method  fuses data from multiple  sensors in one ADNN  at which informative 	features are being extracted and utilized for damage identification.  It generates probabilistic  anomaly scores  to detect damage, asses its severity and further   localize it via a new localization layer introduced in the ADNN. 
		
		We evaluated our method on multi-way datasets in the area of structural health monitoring for damage diagnosis purposes. The data was collected from our deployed data acquisition system on a cable-stayed bridge in Western Sydney and  from a laboratory based  building structure obtained from Los Alamos National Laboratory (LANL). Experimental results show that the proposed method can accurately detect structural damage. It was also able to estimate the different levels of damage severity, and capture damage locations in an  unsupervised aspect. Compared to the state-of-the-art approaches, our proposed method shows better performance in terms of damage	detection and localization. 
	
\end{abstract}

%
%


\maketitle

\renewcommand{\shortauthors}{A. Anaissi et al.}

\subsection{Introduction}

The concept of  smart infrastructure maintenance emerged in  the recent years as a continuous automated process  known as structural health monitoring (SHM). It aims to build a condition-based inspection system driven by data for early damage identification  which results in better life-safety and economic benefits.  Most of the current structural maintenance's approaches are considered as a time-based visual inspection  which often follows a  predefined regular schedule. This kind of time-based inspection for a such structure may results   in certain economic and potential life losses if it was too late or too early. Moreover, some structures such as high bridges raise other challenges in terms  of  accessibility.  SHM has earned a lot of interests during the   last decade due to the fact that it  leads to enhance understanding  the behaviour of infrastructure and increases its life span  whilst maintaining a high level of life-safety.  

\begin{figure}[!h]
	\centering
	\includegraphics[scale=0.35]{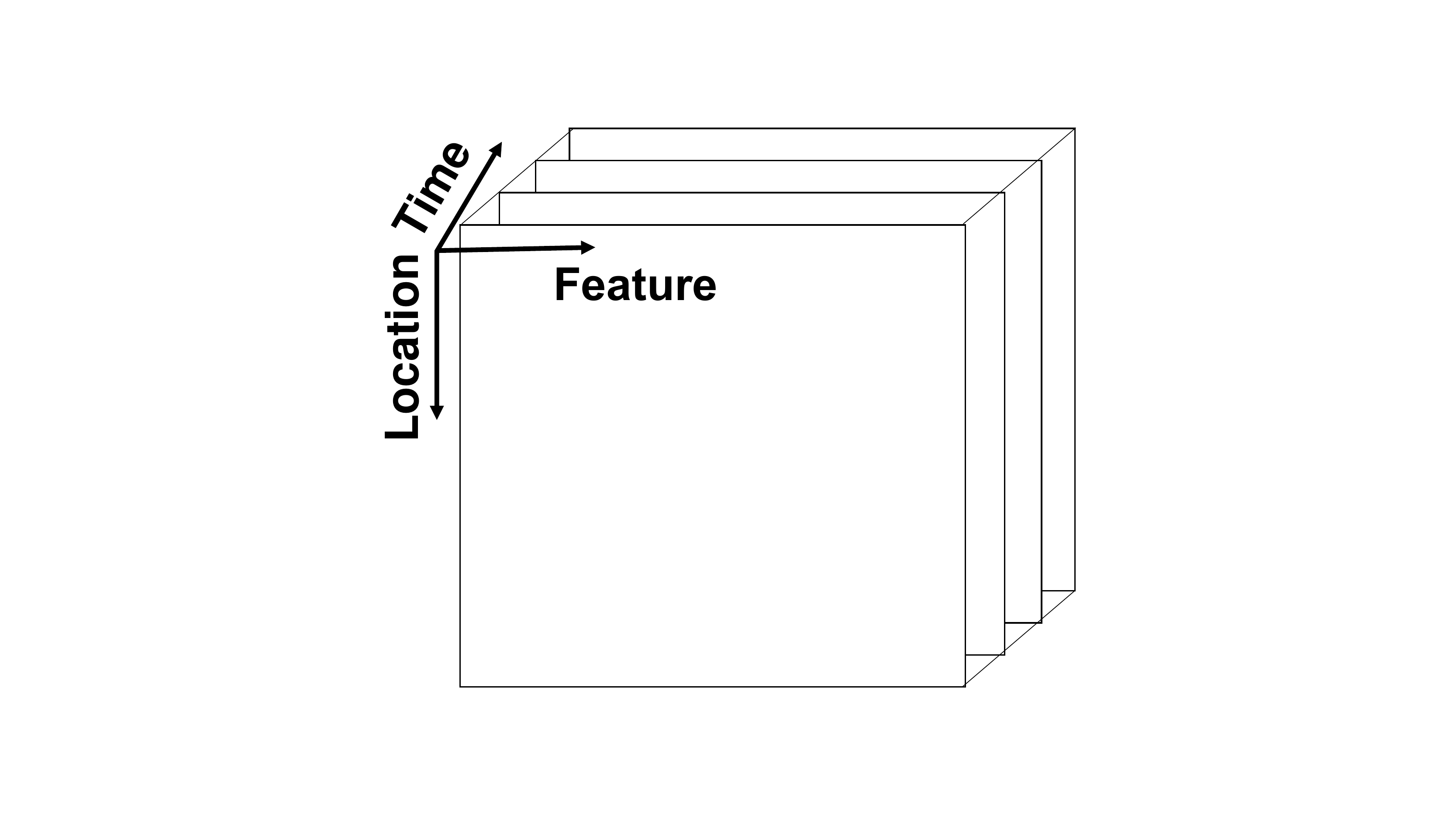}
	\caption{Multi-way data with three modes in SHM applications.}
	\label{f:tensor}
\end{figure}

In the realm of data science,  SHM has  attracted many researchers working in the areas of machine learning and data mining to handle the wealth of vibration responses being simultaneously measured over time  by many sensors attached to a structure   at different locations, and further to identify structural damage. These measured responses lead to high dimensional, multi-way and correlated data which raises many challenges in analyzing and extracting informative features  to learn a damage identification model.  The SHM sensing data  can be arranged  as a three-way data (feature $\times$ location $\times$ time) as described in Figure \ref{f:tensor}. Feature is the information  extracted from the raw signals in time domain (e.g. features in frequency domain). Location represents sensors, and time is data snapshots at different timestamps. Each cell  is a feature value extracted from a particular  sensor at a certain time.

Rytter classified damage identification  into four different levels of complexity \cite{rytter1993vibrational}: damage detection (level 1), localization (level 2), severity assessment (level 3) and failure prediction (level 4). The damage detection level can be solved using two-way analysis techniques by  constructing a standard   anomaly detection model. However damage localization and severity assessment require multi-way data analysis techniques  to  capture the physical meaning of the structure. On the other hand, level 4  is not considered as a machine learning problem since it requires  understanding of the physical characteristics of the damage progression in the structure. These requirements  have motivated us to study the deep neural networks (DNN) as a feature learning method to handle the complexities associated with the multi-way SHM data.  DNN has become popular and attracted many researchers working in the area of data analytic.  It has been successfully applied to solve  complex pattern recognition problems   such as   vision \cite{krizhevsky2011using} and speech \cite{hinton2012deep}. Sutskever et al \cite{sutskever2014sequence}  claim that  DNN often produces   powerful models that achieve high performance in comparison to  other  state-of-the-art machine learning algorithms.  

Generally speaking, data instances from at least two different classes are required for the training stage in DNN. However, in many applications such as SHM \cite{farrar2006introduction}, only data instances from one state (i.e. undamaged or healthy) are available, and the samples from other states (i.e. damaged), if not impossible, are too difficult or costly to acquire. Thus, the classification process becomes as an anomaly detection problem. Anomaly detection methods build a model based on a  given  positive training dataset, and for a new arrived data instance, the model estimates the agreement between the new instance and the trained model. Data instances which do not fit into the trained model are classified as anomaly \cite{anaissi2017self}. 

In the context  of anomaly detection,  an autoencoder deep neural networks  (ADNN)  model may be more practical when only  data from  positive/normal states  are available. It was originally proposed for dimensionality reduction problems. However, it has  proved throughout several applications that it was very capable  to handle the case of one class learning and  solve  anomaly detection problems. Furthermore, it can be also utilized as a data fusion structure which can constructs an internal representation for input data collected from multiple sources and then extracts anomalous sensitive features. Recently, Anaissi and Zandavi \cite{anaissi2019multi} use ADNN to propose a multi-object autoencoder for fault detection and diagnosis in higher-order data based on the  reconstruction error of ADNN.

This paper is an extension of the aforementioned work in \cite{anaissi2019multi}. In combining with the multi-objective autoencoder in \cite{anaissi2019multi}, this paper utilities the  variational autoencoder method  to   propose  a  multi-objective variational autoencoder (MVA) deep neural network   for damage detection, localization and severity assessment. In  contrast to \cite{anaissi2019multi}, MVA performs damage detection  based on reconstruction probability and not reconstruction  error. It performs data fusion  by taking a frontal slice from a multi-way training data. Stochastic  gradient descent  algorithm is then used to   learn reconstructions that are close to its original input slice followed by constructing a sensor identity matrix which used for damage localization. For each new incoming data slice we calculate its  anomaly score   based on reconstruction probability and we use the generated reconstruction errors for damage assessment. The sensor identity matrix is finally utilized to locate the identified damage. 

This work is part of our broader efforts to apply data-driven SHM approaches to real bridges in operation, including the Sydney Harbour Bridge (SHB). We extensively evaluated our proposed method on laboratory-based and real-life structures datasets. The evaluation shows that MVA model has the capability to perform data fusion and extract damage sensitive features which were able to accurately detecting damage. The reconstruction error also   demonstrates the ability to localize the detected damage.   It further reflects the fact that it has the potential to estimate the severity of damage by analyzing the  obtained reconstruction error values. The contributions of this paper are as follows.

\begin{enumerate}
	\item  Sensing multi-way data are fused using ADNN to efficiently extract damage 	sensitive features and then learn reconstruction  of the original input .
	
	\item Damage detection  is accomplished using reconstruction probability which has the capability to identify damage without using any preset fixed threshold parameter. 
	
	\item Damage localization  is accomplished using a new layer introduced  in ADNN.
	
	\item Experiments using data obtained from laboratory-based and real-life structures datasets show the effectiveness of our approach in damage identification and localization.
\end{enumerate}

The remainder of this paper is structured as follows. Section 2 reviews some related work. Section 3 describes our novel MVA method for learning reconstruction error and localizing anomalous data, while Section 4 presents our experimental results and evaluations. Finally, Section 5 discusses the contributions, future work and concludes this paper.

\section{Related Work}
\label{s:related}

Anomaly detection methods have been employed in many application domains such as damage detection in civil structure  \cite{anaissi2018adaptive,anaissi2017damage,anaissi2018automated,khoa2018structural},  intrusion detection in network \cite{leung2005unsupervised,mukkamala2002intrusion} and numerous other fields. They are mainly proposed to handle the cases when only  normal/positive data are available. For instance, \cite{yin2014fault} designed a robust one-class support vector machine (OCSVM)  to eliminate the influence of outliers to the learned boundary and used it to detect damage in a simulated structure. Mahadevan and Shin \textit{et~al.} \cite{yin2014fault} and \cite{mahadevan2009fault}, proposed an approach for fault detection and diagnosis using OCSVM and SVM-recursive feature elimination. Further, the authors \cite{yin2014fault} and \cite{mahadevan2009fault} used OCSVM to detect damage in a rotating machinery and the results showed that the performance of the proposed method is superior to the state-of-the art methods. However, the work above focused on damage detection  using  two-way matrix data generated via individual sensor which might help in detecting  damage  but not in assessing its  severity or localize it. 

In the recent years, various data fusion methods have been used in SHM applications to deal with the multi-way data \cite{sophian2003feature,lu2009feature,anaissi2018tensor}. Some of these methods  performed data fusion in an unsophisticated manner by simply concatenating features obtained from different sensors \cite{sophian2003feature}. However, more advanced methods including principle component analysis (PCA), neural networks and Bayesian methods have been adopted at this level \cite{jiang2006structural}. In this context, khoa \textit{ et al.} \cite{khoa2017smart} used advanced tensor  analysis  to fuse data from multiple sensors followed by constructing  a OCSVM model for damage detection. The authors were able to successfully detect and assess the severity of the damage but failed to localize it.    

With the advent of deep learning methods, ADNN attracted many researchers working in the area of anomaly detection due its promising achievements in many domains \cite{hong2015multimodal,ap2014autoencoder,germain2015made}. Jinwon and Sungzoon \cite{an2015variational} propose   a variational autoencoders  for anomaly detection tasks. They used a probability measure to generate the anomaly score instead of reconstruction error. The work in \cite{chong2017abnormal} also uses   autoencoders   for anomaly detection in videos.  The authors evaluate their method on real-world datasets and  reported better performance    over   other state-of-the-art methods.  The authors in \cite{yan2015accurate}  use  deep learning methods to hierarchically learn features from the sensor measurements of exhaust gas temperatures. Then they use the learned features as  input to an ADNN  for performing combustor anomaly detection. 

In fact, there are still few works in which researchers try to apply ADNN methods  to other data analytic tasks such as data fusion in  multi-way datasets. In this study, we propose a MVA  deep  neural network as a data fusion method to extract  damage sensitive features from three-way  measured  responses and to perform damage detection based on  the reconstruction probability. Further, the average distance between the anomaly scores of each corresponding sensor nodes  are used as an another measure to localize  and assess the severity of structural damage.

\section{Background}
\subsection{Autoencoder Deep Neural Network}
\label{s:Autoencoder}

\begin{figure}
	\centering
	\includegraphics[trim=2cm 3cm 8cm 2cm,clip=true,scale=0.5]{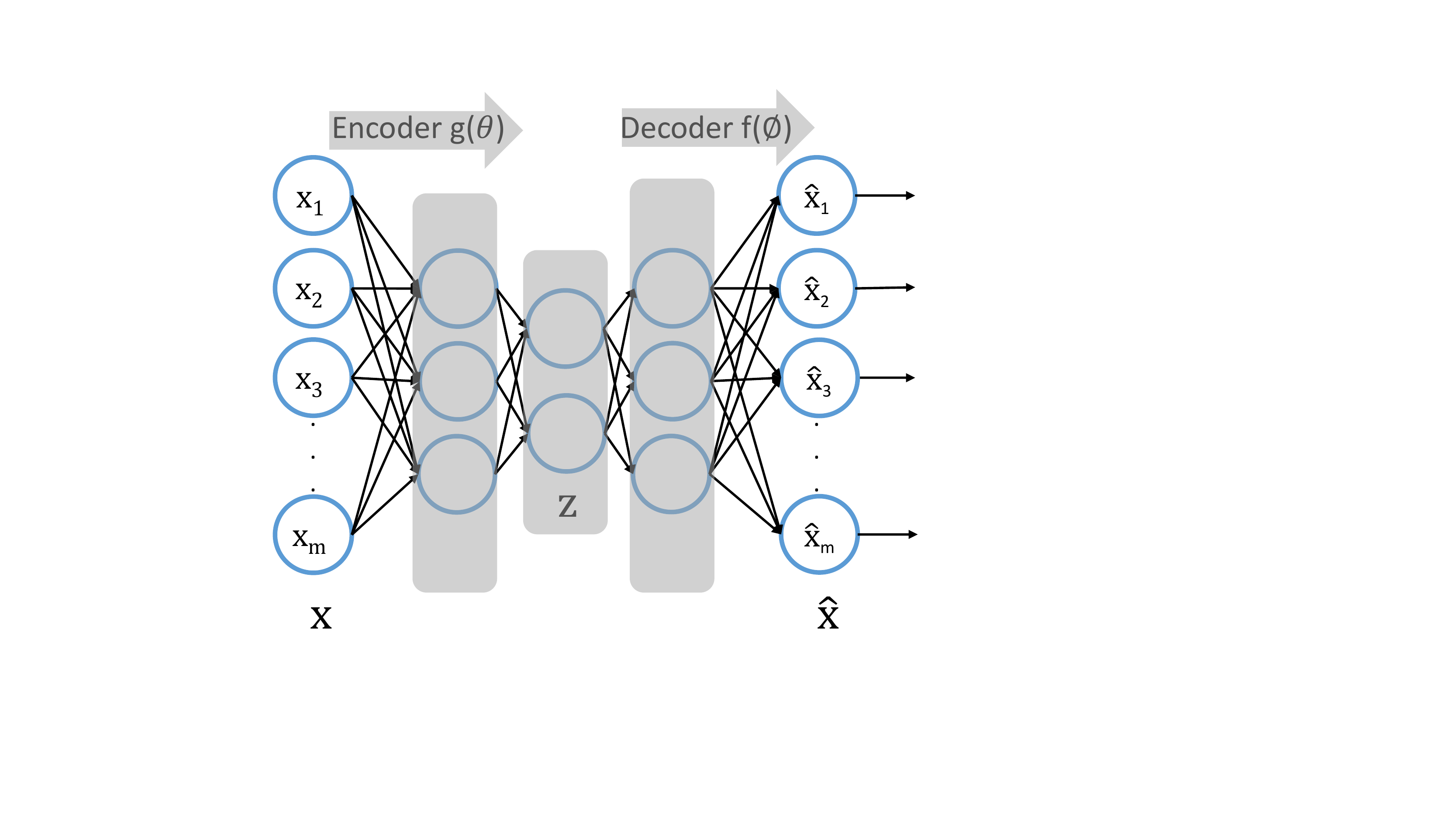}
	\caption{Autoencoder  neural network architecture.}
	\label{f:ae}
\end{figure}

Autoencoder deep neural network is an unsupervised learning process which has the ability to learn from one class data. It is an extension to the deep neural network which basically designed for supervised learning when the class labels are given with the training examples. The rational  idea of  an autoencoder  is to  force the network  to  learn a  lower dimensional space $Z$ for the input features $X$, and then try  to reconstruct the original feature space to $\hat{X}$. In other words, it  sits the target values to be approximately equal to its original inputs. In this sense, the main  objective of autoencoders is to learn  reproducing  input vectors $\{x_1, x_2, x_3, \dots, x_m\}$ as outputs $\{\hat{x}_1, \hat{x}_2, \hat{x}_3, \dots, \hat{x}_m\}$.  Figure \ref{ae} illustrates the architecture of  ADNN   composed of $L$ hidden  layers  ($L=3$ for simplification).   Layer $X$ is the input layer which encoded into the middle layer  $Z$, and then decoded into the output layer $\hat{X}$. Each layer  consists  from a set of  nodes denoted by circle in Figure \ref{f:ae}. The  nodes in the input   layer represents the input features which are often aligned with the number of  features for a given dataset. However, the number of nodes in the hidden layer(s) are selected by  user. In contrast to the traditional neural network, the number of nodes in the output layer are aligned with the same  number of the input layers.

The learning process of ADNN  successively  computes the  output  of each node in the network. For a node $i$ in layer $l$ we calculate an output value  $z^{(l)}_i$ obtained  by  computing the  total weighted $W_{ij}$  sum of  the input values  plus the  bias term $b_i$ using the following equation:

\begin{eqnarray}
\label{z}
z_i^{(l)} = \sum_{j=1}^{n}W_{ij}^{(l-1)} a_j^{(l-1)} + b_i^{(l)} 
\end{eqnarray}

The parameter $W$ is the coefficient weight  written as $W_{ij}$  when associated with the connection between node $j$ in layer $l-1$, and node $i$ in layer $l$.  The  $b_i$ parameter is the bias term associated with the node $i$ in layer $l$ and $a_j^{(l-1)}$ is the output value of node $j$ in layer $l-1$. The resultant output is then processed through  an activation function denoted by $a^{(l)}_i$, and it is defined as follows:
\begin{eqnarray}
\label{a}
a_i^{(l)} =  f(z_i^{(l)})
\end{eqnarray}

Intuitively,  in the input layer $a^1 = x$, and in the   output layer,  $a^3 = \hat{x}$. The most common activation functions in the hidden layers are  the sigmoid and hyperbolic tangent  defined in Equations \ref{sig} and \ref{tanh}, respectively. However, in autoencoder settings a linear function is used in the output layer since we don't scale the output of the network to a specific interval  $([0, 1]$ or $[-1, 1])$.

\begin{eqnarray}
\label{sig}
f(z) = \frac{1}{1 + e^{-z}}
\end{eqnarray}

\begin{eqnarray}
\label{tanh}
f(z) = \frac{ e^{z} - e^{-z} }{e^{z} + e^{-z}}
\end{eqnarray}

Lets say that an autoencoder is composed of two systems known as  encoder $g(\theta)$ and decoder $f(\phi)$. The encoder maps an input vector $X$ to a latent vector $Z$. Then the decoder maps $Z$ back to the original input feature $\hat{X}$. The autoencoder uses back propagation algorithm to learn the  parameters $(\theta, \phi)$. In each iteration of the training process,  we perform a feedforward pass which successively computes the output values $a_i^{(l)}$ for all layer's nodes.  Once completed, we calculate the  cost error $J(\theta, \phi)$  using Equation \ref{cost}  and then propagate it backward to the network layer.
 
 \begin{align}
 \label{cost}
 J(\theta, \phi) &=\frac{1}{n} \sum_{i=1}^{n} \Bigg( \frac{1}{2} \Vert x^{(i)}  - \hat{x}^{(i)}\Vert^2 \Bigg) \\ \nonumber
 &=\frac{1}{n} \sum_{i=1}^{n} \Bigg( \frac{1}{n} \Vert x^{(i)}  - f_{\theta} (g_{\phi}( \hat{x}^{(i)}))\Vert^2 \Bigg) 
 \end{align}
 
 In this setting,  we perform a stochastic gradient descent step to update the learning parameters $(\theta, \phi)$.  This is done by computing the partial derivative of the cost function $J(\theta, \phi)$ (defined in Equation \ref{cost}) with respect to $\theta$ and $\phi$  as follows:

\begin{eqnarray}
\label{primew}
\theta  := \theta  - \alpha  \frac{\partial }{\partial \theta }  J(\theta, \phi)
\end{eqnarray}

We update $\phi$ in the same way.  The complete steps are summarized in  Algorithm \ref{ae}.

 \begin{algorithm}[h!]
 	\caption{Autoencoder training algorithm}
 	\label{ae}
 	\textbf{Input}: A set of $n$  positive samples $x=\{{x^{(i)}}\}_{i=1}^n$\\
 	\begin{enumerate}
 		\item[$\bullet$] Initialize  $(\theta,\phi)$ to a small random value  $\mathcal{N}(0, \epsilon^2)$
 	\end{enumerate}
   \textbf{repeat} 
 	\textbf{For} $i=1:n$	
 	\begin{enumerate}
 		
 		\item[$\bullet$]    Perform a  feedforward pass to compute  all nodes activations using Equations \ref{z} and  \ref{a}.
 		
 		\item[$\bullet$] Compute the sum of reconstruction  cost error   $J(\theta, \phi)$ using Equation \ref{cost}
 		
 		\item[$\bullet$] Update  $(\theta,\phi)$ using Equation \ref{primew}
 		
 	\end{enumerate}	
 	\textbf{until} convergence of parameters $\theta, \phi$
 	
 	\textbf{Output}: encoder $f(\theta)$, decoder $g(\phi)$.
 \end{algorithm}

 Once the autoencoder get trained, the network will be able to reconstruct an new incoming positive data, while it fails with anomalous data. This will be judged based on the reconstruction error (RE) which is measured by applying the  Euclidean norm to the difference between the input and output nodes as shown in Equation \ref{re}.

\begin{eqnarray}
	\label{re}
	RE(x) = \Vert x^{(i)} - \hat{x}^{(i)}\Vert^2 
\end{eqnarray}

The measured value of  RE is used as anomaly score for a given new sample. Intuitively, examples from the similar distribution to the training data should have low reconstruction error, whereas anomalies should have high anomaly score.  Algorithm \ref{ad} shows the process of anomaly detection based on the reconstruction error of autoencoders.

\begin{algorithm}[h!]
	\caption{Autoencoder anomaly detection algorithm}
	\label{ad}
	\textbf{Input}: A set of  new arrived samples $x=\{{x^{(i)}}\}_{i=1}^n$, and $\alpha$ \\
	\begin{enumerate}
		\item[$\bullet$] $\theta, \phi \leftarrow$  Algorithm \ref{ae} 
	\end{enumerate}
	\textbf{For} $i=1:n$	
	\begin{enumerate}		
		\item[$\bullet$] Perform a  feedforward pass to compute  all nodes' activations using Equations \ref{z} and  \ref{a}.		
		\item[$\bullet$] Compute $RE$ using Equation \ref{re}
		
		\item[$\bullet$] \textbf{if} $RE  > \alpha$ 
			\begin{enumerate}
				\item[$\bullet$]  $x^{(i)}$  is an anomaly
			\end{enumerate}
	  \item[$\bullet$] \textbf{else}
		\begin{enumerate}
			\item[$\bullet$]  $x^{(i)}$ is normal
		\end{enumerate}		
	\end{enumerate}	
\end{algorithm}

\section{Multi-Objective Variational Autoencoder}
\label{s:method}

\begin{figure}
	\centering
	\includegraphics[scale=0.4]{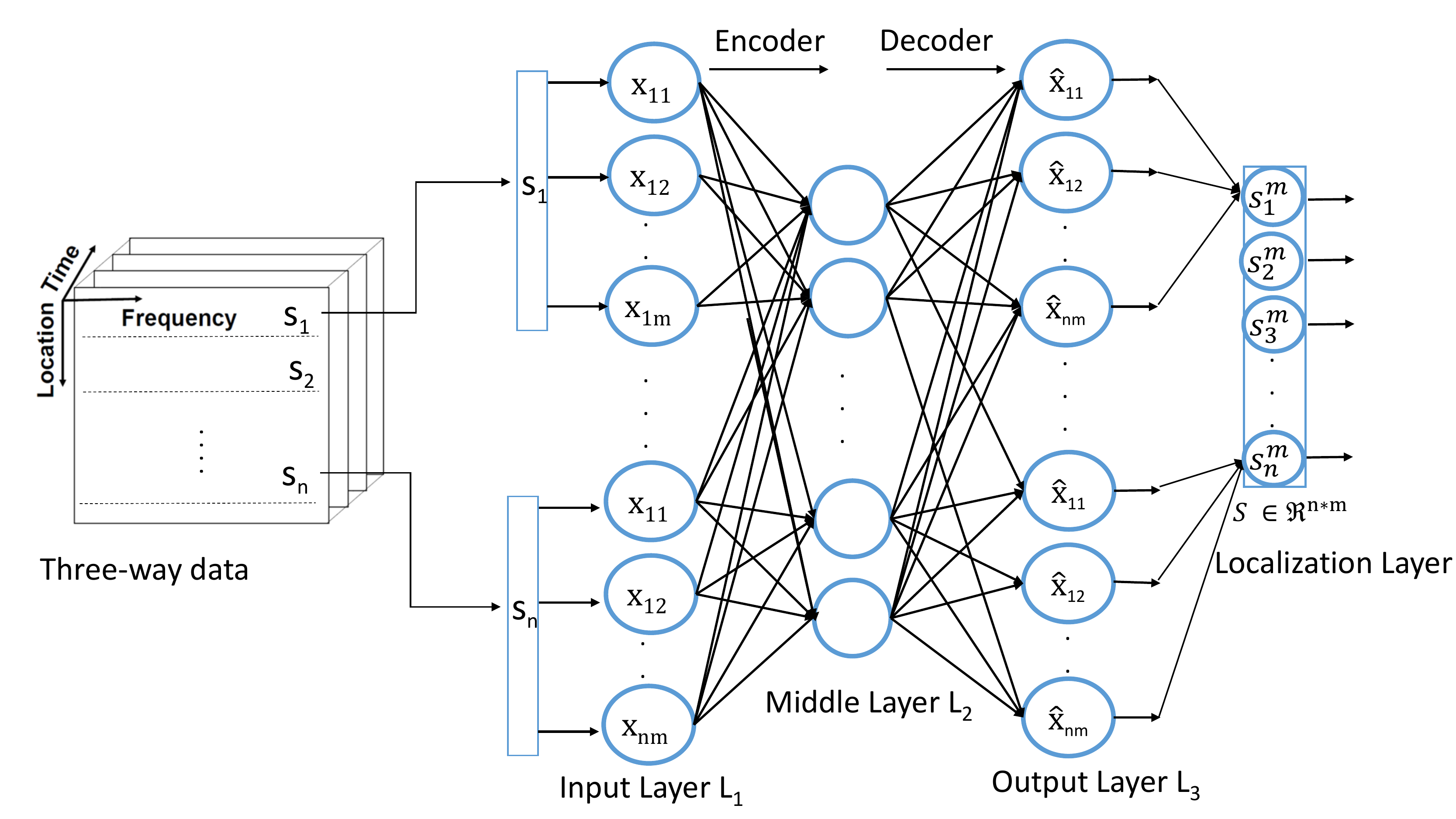}
	\caption{Autoencoder deep neural network architecture  of MVA.}
	\label{f:frame}
\end{figure}
 
We propose  a multi-objective variational autoencoder (MVA) neural network   for damage detection and diagnosis based on the reconstruction probability  of ADNN.  Our MVA method performs multi-way data fusion  by taking a frontal slice from the training  data (as shown in Figure \ref{f:frame}). Each input slice represents all feature signals across all locations at a particular  time. Stochastic  gradient descent  algorithm is used here  to learn reconstructions that are close to its original input slice. Once the network get trained, we  create a sensor  identity matrix $ S \in \Re^{s \times m}$ in which each row captures meaningful information for each sensor location for damage localization purposes.  The values in this matrix are obtained  by calculating  the average  total reconstruction error for  each set  of $m$ output nodes related to one single sensor. 

Our  method employed the concept of variational auto encoder (VAE)   for computing the anomaly score for each new incoming data slice.  It aims to  calculate the   anomaly score for new arrived data  based on its reconstruction probability. This  measure provides more principled and objective decision value than reconstruction errors since it considers the variability of the distribution variables, and does not require presetting fixed threshold parameter for identifying  damage.  Setting a threshold for reconstruction error is problematic especially in the case of muti-way heterogeneous data. Moreover,  the normal  and anomaly data might shares the same mean value. However, anomalous data will not share the same variance to the normal data and it leads to significant  lower reconstruction probability, thus classified as damage. The following sections discuss the details of the  proposed method.

\subsection{  Multi-Way Data Fusion}

As we  observed in this study,  a large number of sensors are usually used to collect data  in SHM applications which often aim to monitor large civil structures such as bridge or a high-rise building. The sensing data being generated from networked sensors mounted structures are considered as three-way data in the form  of ($location \times frequency \times time$) as previously described in Figure \ref{f:tensor}.  In this setting, two-way matrix analysis is not able to capture the correlation between sensors \cite{acar2009}. At the same time, unfolding the three-way data and concatenating the frequency features from multiple sensors at a certain time to form a single data instance at that time  may result in information loss  since it breaks the modular structure inherent in  three-way  data ~\cite{acar2009}.  Accordingly, data fusion plays a critical role in analyzing  structure behaviours   and assessing the severity of any damage data.  

Basically, ADNN is mainly used for the purpose of  dimensionality reduction  or as anomaly detection models. In fact, ADNN can be also utilized as data fusion structure which can constructs an internal representation for  input data collected from multiple sources i.e. sensors. Therefore, our MVA method utilizes the ADNN as a multi-way data fusion  model which automatically  learns features via its deep-layered structure. 

As shown in Figure \ref{f:frame},   ADNN model receives data from multiple sensors at the same time  by taking a frontal slice from a training three-way data. Each input slice represents all feature signals across all locations at a particular  time. This data from multiple sensors is fed  into the input layer  to extract damage sensitive features via the  encoder layers. The resultant new features in the middle layer ($Z$) are then used   by the decoder layers to determine  the  damage detection results.

\subsection{ Probabilistic Anomaly Detection} 
The rational idea of anomaly detection in ADNN is to see how well a new data point follows the normal examples. We mentioned before that ADNN aims to  learn (encoder) a  lower dimensional space $Z$ for  input features $X$, and then try  to reconstruct (decode) the original feature space $\hat{X}$. Let's denote the encoder and decoder by  $q_{\phi}(Z \mid X)$ and $p_{\theta}(X \mid Z)$, respectively. This representation  is known as the conditional probability. For example, $p_{\theta}(X \mid Z)$ is the conditional of $X$ such that $Z$ has happened. Intuitively, the decoder process yields to information loss because the data goes from a low dimensional space $Z$   to a larger dimensional space $\hat{X}$. This loss is known as the reconstruction error which can be measured  by calculating the  log-likelihood $\log p_{\theta}(X \mid Z)$ and it will be eventually used as an anomaly score. This measure allows us  to see how effectively the decoder has learned to reconstruct an input features $X$ given its latent representation $Z$. 

Our probabilistic anomaly detection method follows the concept of VAE   to find a distribution  of some latent variable $Z$ which we can sample from $Z \sim q_{\phi}(Z \mid X)$  to generate new samples   $\hat{X}$ from $ p_{\theta}(X \mid Z)$. Each latent variable $z_i$ represents   a probability distribution for a given input feature. In the  decoding process, we randomly sample from this  latent state distribution to generate a vector to be used as an input for the decoder model.

Given $X$ be a set of observed variables and $Z$  is the set of latent variables, the objective function of VAE is considered as an inference problem which aims to compute the conditional distribution of latent variables $Z$  given the observations $X$ i.e.  $p(Z \mid X)$. Using  Bayesian theorem, we can write it as follows:
\begin{eqnarray}
\label{eq:bayess}
p_{\theta}(Z \mid X)  = \frac{P_{\theta}(X \mid Z)  \times P(Z)}{P(X)}
\end{eqnarray}
However, calculating  the evidence $p(X)$ is not practical since it  requires computing a multidimensional integral in the $d$ unknown variables $z_1,\dots,z_d$ \cite{kingma2013auto}. Thus, the  variational inference (VI) tool is used here to    perform approximate Bayesian of the  posterior distribution $p_{\theta}(Z \mid X)$  with a  parametric family of distributions $Q_{\phi}(Z \mid X)$ in a such  way that it has tractable solution. The main idea of VI is to pose the inference problem as an optimization problem by modeling $p(Z \mid X)$  using $Q(Z \mid X)$ where  $Q(Z \mid X)$ has a simple distribution such as Gaussian. 

The $\mathcal{KL}$ divergence method defined in Equation \ref{eq:kl1} is   used here to measure the information loss between the two probability distributions  $p(Z \mid X)$ and $Q(Z \mid X)$. In this sense, the optimization problem is to minimize the $\mathcal{KL}$ divergence denoted by $\mathcal{D}_{\mathcal{KL}}$ i.e. ($\min_{\mathcal{KL}} \;\;p(Z \mid X) \;\; || \;\;  Q(Z \mid X))$).

\begin{align}
\label{eq:kl1} 
\mathcal{D}_{\mathcal{KL}} (p_{\theta}(Z \mid X) \;\; || \;\;  Q_{\phi}(Z \mid X)) 
                        &= \sum_{z}  Q_{\phi}(Z \mid X) \log (\frac{Q_{\phi}(Z \mid X)}{p_{\theta}(Z \mid X)})\\ \nonumber
						&= E_{Z \sim Q_{\phi}(Z \mid X)} \big[ \log \frac{Q_{\phi}(Z \mid X)}{p_{\theta}(Z \mid X)} \big]\\ \nonumber
						&= E_{Z \sim Q_{\phi}(Z \mid X)} \big[ \log(Q_{\phi}(Z \mid X)) - \log(p_{\theta}(Z \mid X)) \big]
\end{align}

By substituting Equations \ref{eq:bayess} in   \ref{eq:kl1}, the resultant equation will be as follows:
 
 \begin{align}
 \label{eq:kl2} 
 \mathcal{D}_{\mathcal{KL}} (p_{\theta}(Z \mid X) \;\; || \;\;  Q_{\phi}(Z \mid X)) 
  &=E_{Z} \big[ \log(Q_{\phi}(Z \mid X)) - \log \frac{P_{\theta}(X \mid Z)  \times P_{\theta}(Z)}{P_{\theta}(X)} \big]\\ \nonumber 
  &=E_{Z} \big[ \log(Q_{\phi}(Z \mid X)) - \log P_{\theta}(X \mid Z) -\log P_{\theta}(Z)  + \log P_{\theta}(X) \big] 
 \end{align}
 where $Z = Z \sim Q_{\phi}(Z \mid X)$.  Since the the expectation (E) is based on $Z$ and $P_{\theta}(X)$ does not involve $Z$, we can remove $P_{\theta}(X)$ from Equation  \ref{eq:kl2} and write it  as follows:
 \begin{align}
\label{eq:kl3} 
\log P_{\theta}(X) - \mathcal{D}_{\mathcal{KL}} (p_{\theta}(Z \mid X)  || Q_{\phi}(Z \mid X)) &= E_{Z} \big[ \log(p_{\theta}(X \mid Z))\big] -E_{Z} \big[ \log(Q_{\phi}(Z \mid X)) -\log P_{\theta}(Z) \big]
\end{align} 
The final objective function of variational  autoencoder is as follows:
 \begin{align}
\label{eq:obj} 
\log P_{\theta}(X) - \mathcal{D}_{\mathcal{KL}} (p_{\theta}(Z \mid X)  || Q_{\phi}(Z \mid X)) = E_{Z} \big[ \log(p_{\theta}(X \mid Z))\big] - \mathcal{D}_{\mathcal{KL}}  Q_{\phi}(Z \mid X)  || p_{\theta}(Z)
\end{align}  
The first term i.e. $\log(p_{\theta}(X \mid Z))$  represents the reconstruction likelihood and the second term i.e $\mathcal{D}_{\mathcal{KL}}$ is the regularization parameter which forces the posterior distribution  $Q_{\phi}(Z \mid X)$  to be similar to the prior distribution $p_{\theta}(Z)$. The loss function$J(\theta,\phi)$ of our autoencoder is the negative value of the objective function and its defined as:
  \begin{align}
 \label{eq:loss} 
 J(\theta,\phi) = - E_{Z} \big[ \log(p_{\theta}(X \mid Z))\big] + \mathcal{D}_{\mathcal{KL}}  \big[ Q_{\phi}(Z \mid X)  || p_{\theta}(Z) \big]
 \end{align} 
In variational Bayesian method, this loss function is known as the variational lower bound or evidence lower bound (ELBO). This "lower bound" part comes from the fact that $\mathcal{KL}$ divergence is always non-negative. Thus $J(\theta,\phi)$ is the lower bound of $\log P_{\theta}(X)$, and it is also known  that  $ \mathcal{D}_{\mathcal{KL}}  \big[ q_{\phi}(Z \mid X, \lambda)  || p_{\theta}(Z  \mid X) \big] \geq = 0$. As a result $J(\theta,\phi) \le \log P_{\theta}(X) $. Therefore  by minimizing the  loss, we are maximizing the lower bound of the probability generating real data samples.

Now we need to train the variational autoencoder to learn $Q_{\phi}(Z \mid X)$ using gradient descent algorithm to optimize the loss with respect to the parameters  $\theta, \phi$ . This is where the VAE can relate to the autoencoder where the encoder model learns $Q_{\phi}(Z \mid X)$ by  mapping $X$  to $Z$ and the decoder model learns $p_{\theta}(Z \mid X)$ by  mapping $Z$ back to $X$. For stochastic gradient descent with step size $\alpha$, the encoder parameters are updated using  Equation \ref{primew}. Once $Q_{\phi}(Z \mid X)$ is learned, we sample the latent vector $Z$ from $q_{\phi}(Z \mid X)$ and then feed it into the decoder network  $p_{\theta}(X \mid Z)$ to generate  the new data $\hat{X}$. The training steps of MVA are illustrated in  Algorithm \ref{VAE}.

  \begin{algorithm}[h!]
	\caption{MVA training algorithm  }
	\label{VAE}
	\textbf{Input}: A set of $n$  positive samples $x=\{{x^{(i)}}\}_{i=1}^n$\\
	\begin{enumerate}
			\item[$\bullet$] Initialize  $(\phi,\theta)$ to a small random value  $\mathcal{N}(0, \epsilon^2)$
	\end{enumerate}

    \textbf{repeat} 
    \begin{enumerate}
	  \item[$\bullet$] \textbf{for} $i=1:n$	
		\begin{enumerate}
			\item[]
			\begin{enumerate}		
			\item[$\bullet$] Generate $L$ samples from $\epsilon \sim \mathcal{N}(0,1)	$
		    \item[$\bullet$] Perform a  feedforward pass to compute  all nodes' activations.	
		\end{enumerate}
	\end{enumerate}
	
	\end{enumerate}
	\begin{enumerate}
	  \item[$\bullet$]\textbf{end for} \\

    \item[$\bullet$]  Compute the error usig Equation \ref{eq:loss} \\
     \item[$\bullet$]  Update  $(\phi,\theta)$ using gradients of E  
    \end{enumerate}
     \textbf{until} convergence of parameters  $(\phi,\theta)$
\end{algorithm}

To get the reconstruction $\hat{X}$, we  generate $L$ random samples from $ z \sim N (\mu_{{z}^{(i)}}, \sigma_{{z}^{(i)}})$ where $\mu_{{z}^{(i)}}$ and $\sigma{{z}^{(i)}}$ are the mean and standard deviation of the middle layer $z|x^i$  in ADNN, respectively. For each random sample in $L$, we calculate $\mu_{\hat{x}^{(i)}}$ and $\sigma_{\hat{x}^{(i,l)}}$ for the output layer in  ADNN. The final reconstruction probability (RB) can be estimated as follows:

\begin{eqnarray}    
\label{rb}
RP(x_{new}) = \frac{1}{L} \sum_{l=1}^{L} p_{\theta}(x \mid z^{(i,l)}|\mu_{\hat{x}^{(i,l)}},\sigma_{\hat{x}^{(i,l)}})
\end{eqnarray}	
The damage detection steps of MVA are illustrated in  Algorithm \ref{VAE-d}.
  \begin{algorithm}[h!]
	\caption{MVA damage detection algorithm}
	\label{VAE-d}
	\textbf{Input}: A set of  new arrived samples ${X}_{i=1}^n \in \Re^{n \times m \times s}$\\
	\begin{enumerate}
		\item[$\bullet$] $(\phi,\theta)$ $\leftarrow$  Algorithm \ref{VAE} 
	\end{enumerate}
	\textbf{For} $i=1:n$	
	\begin{enumerate}		
		\item[$\bullet$] $\mu_{{z}^{(i)}}, \sigma_{{z}^{(i)}} = q_{\phi}(z|x^{(i)})$		
		\item[$\bullet$] Generate $L$ samples from $ z \sim N (\mu_{{z}^{(i)}}, \sigma_{{z}^{(i)}})$
		
		\item[$\bullet$]\textbf{For} $l=1:L$
		\begin{enumerate}
			\item[$\bullet$] $\mu_{\hat{x}^{(i,l)}}, \sigma_{\hat{x}^{(i,l)}} = p_{\theta}(x|z^{(i,l)})$		
		\end{enumerate}
	    \item[$\bullet$]\textbf{end for} \\	
	
		 \item[$\bullet$]Compute $RB^{(i)}$ using Equation \ref{rb}
		 \item[$\bullet$]  \textbf{if} reconstruction probability(i) $ \le \alpha$ \textbf{then}
		 \begin{enumerate}
		 	 \item[$\bullet$]$x^{(i)}$ is  healthy		 
     	\end{enumerate}	
       \item[$\bullet$] \textbf{else}
        \begin{enumerate}
       	 \item[$\bullet$]$x^{(i)}$ is  damage		 
       \end{enumerate}
   	
    \end{enumerate}
    \textbf{end for} 
\end{algorithm}

\subsection{Damage Localization}

Once a new data slice  identified  as anomaly by  ADNN, the  values from the output nodes are further propagated  into another layer called localization layer as illustrated in Figure \ref{f:frame}. It   consists from a set of  $n$ nodes each  representing one sensor data source. The purpose of this layer is to solve the problem of fault localization. The output values to this layer are obtained  by calculating  the average of the total reconstruction error for  each $m$   output nodes related to  one sensor.   The resultant outputs are stored in a matrix $S \in \Re^{n \times m}$ where $n$ is the number of sensors and $m$ is the number of features for each sensor.  Using  $S$ matrix, it is possible to perform a $k$-nearest neighbouring algorithm on new  output scores $S_{new}$ with each  row of  matrix $S$ to locate the anomalous rows. The average distance difference between  $S$ and $S_{new}$ is used as another anomaly score for damage localization.

\section{Experimental Results}
\label{s:results}

\subsection{Data Collection}
We conducted experiments on two case studies representing typical
types of civil structures. The first case study is a real data collected from a  cable-stayed bridge in  Western Sydney, Australia. The second one is  a laboratory based  building structure obtained from Los Alamos National Laboratory (LANL) \cite{larson1987alamos}. 
\subsubsection{The Cable-Stayed Bridge}
\label{s:data_wsu}

\begin{figure*}
	\centering
	\includegraphics[width=\textwidth]{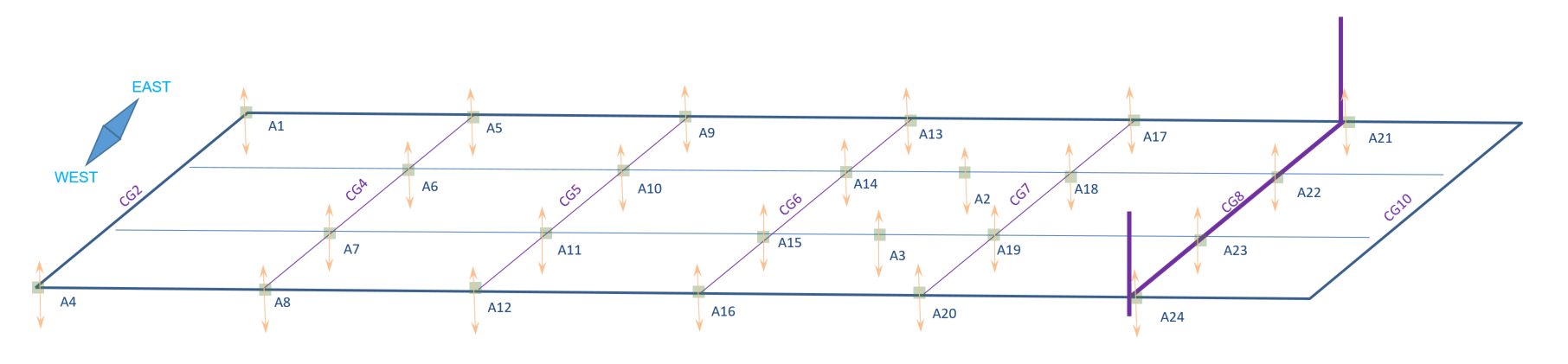}
	\caption{The locations on the bridge's deck of the 24 $Ai$ accelerometers used in this study. The cross girder $j$ of the bridge is displayed as $CGj$.}
	\label{fig:wsuloc}
\end{figure*}

The bridge was instrumented by 24 uniaxial accelerometers and 28 strain gauges.
 In this paper we are using only features based on accelerations data collected from sensors $Ai$ with $i\in [1;24]$.   Figure~\ref{fig:wsuloc} shows the locations of these 24 sensors  on the bridge deck.  

For the sake of experiments, we emulated two different kind of damage on this bridge by placing a large static load (vehicle) at different location of a structure. Thus, three scenarios have been considered which includes: no vehicle is placed on the bridge (healthy state), a light vehicle with approximate mass of 3 t is placed on the bridge close to  location A10 ("Car-Damage") and a bus with approximate mass of 12.5 t is located on the bridge at location A14 ("Bus-Damage").  This emulates slight and severe damage cases  which were used in our evaluation Section~\ref{s:wsueval}.  



\subsubsection{Building Data}
\label{s:data_b}
\begin{figure}
	\centering
	\includegraphics[scale=0.5]{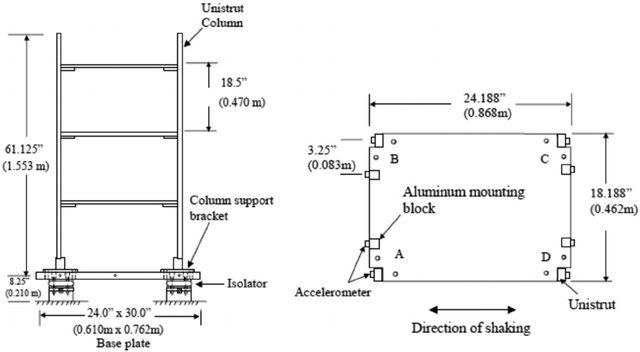}
	\caption{Three-story building and floor layout \cite{larson1987alamos}.}
	\label{fig:alamos}
\end{figure}

Our second case study was based on the a data collected by \cite{larson1987alamos} from three-story building structure. It is  made up of Unistrut columns and aluminum floor plates connected by bolts and brackets as presented in Figure \ref{fig:alamos}. Eight accelerometers were instrumented on each floor (two on each joint). A shaker was placed  at corner D  to generate excitation data. It generates  240 samples (a.k.a. events) separated into two main groups, Healthy (150 samples) and Damaged (90 samples). Each event consists of acceleration data for a period of  5.12 seconds  sampled  at 1600 Hz, resulting in a vector of 8192 frequency values. The Damaged samples  were further partitioned into two different damaged cases based on their location: damage in location 3C (60 samples), and the damage in both locations 1A and 3C (30 samples). The damage was introduced by detaching or loosening the  bolts at the joints, allowing the aluminum floor plate to move freely relative to the Unistrut column.

\subsection{Results and Discussions}
\label{chapter:experiments}

This section demonstrates how our MVA method can successfully detect and assess the severity of structural damage, and further localize it. It is using the sensor-based data from the two case studies described in  Section \ref{s:data_wsu}.

For all experiments, six hidden layers were used in MVA and  the accuracy values were obtained using the F-Score (FS) measure  defined as $\textrm{\Fscore} = 2 \cdot \dfrac{\textrm{Precision}  \times \textrm{Recall} }{\textrm{Precision} + \textrm{Recall}}$ where $\textrm{Precision} = \dfrac{\textrm{TP} }{\textrm{TP} + \textrm{FP}}$ and $\textrm{Recall}  = \dfrac{\textrm{TP} }{\textrm{TP} + \textrm{FN}}$ (the number of true positive, false positive and false negative are abbreviated by TP, FP and FN, respectively).
The OCSVM model was used in these experiments  as a state-of-the-art method for comparison purposes. The rate of anomalies $\nu$ in OCSVM was set to 0.05 and the Gaussian kernel parameter  $\sigma$ was tuned using  a technique proposed by \cite{anaissi2018gaussian}.

\subsubsection{The Cable-Stayed Bridge}
\label{s:wsueval}

Our MVA method was initially validated using vibration data collected from the cable-stayed bridge described in Section \ref{s:data_wsu}. We used 24 uni-axial accelerometers to generate 262 samples (a.k.a events) each  consists of acceleration data for a period of 2 seconds at a sampling rate of 600 Hz. 

For each reading of the uni-axial accelerometer, we  normalized its  magnitude   to have a zero mean and one standard variation. The fast Fourier transform (FFT) is then used to represent the generated data in the frequency
domain.  Each event now  has   a feature   vector of  600 attributes representing  its frequencies. The resultant three-way  data  has a structure of 24 sensors $\times$ 600 features  $\times$ 262 events. We separated the 262 data instances  into two groups,   125 samples related to the healthy state  and 137 samples for damage state. The 137 damage examples were further divided into two different damaged cases: the "Car-Damage" samples (107)   generated when a stationary car was placed on the bridge, and the "Bus-Damage" samples (30) emulated by the stationary bus. 

\begin{figure}
	\centering
	\includegraphics[scale=0.5]{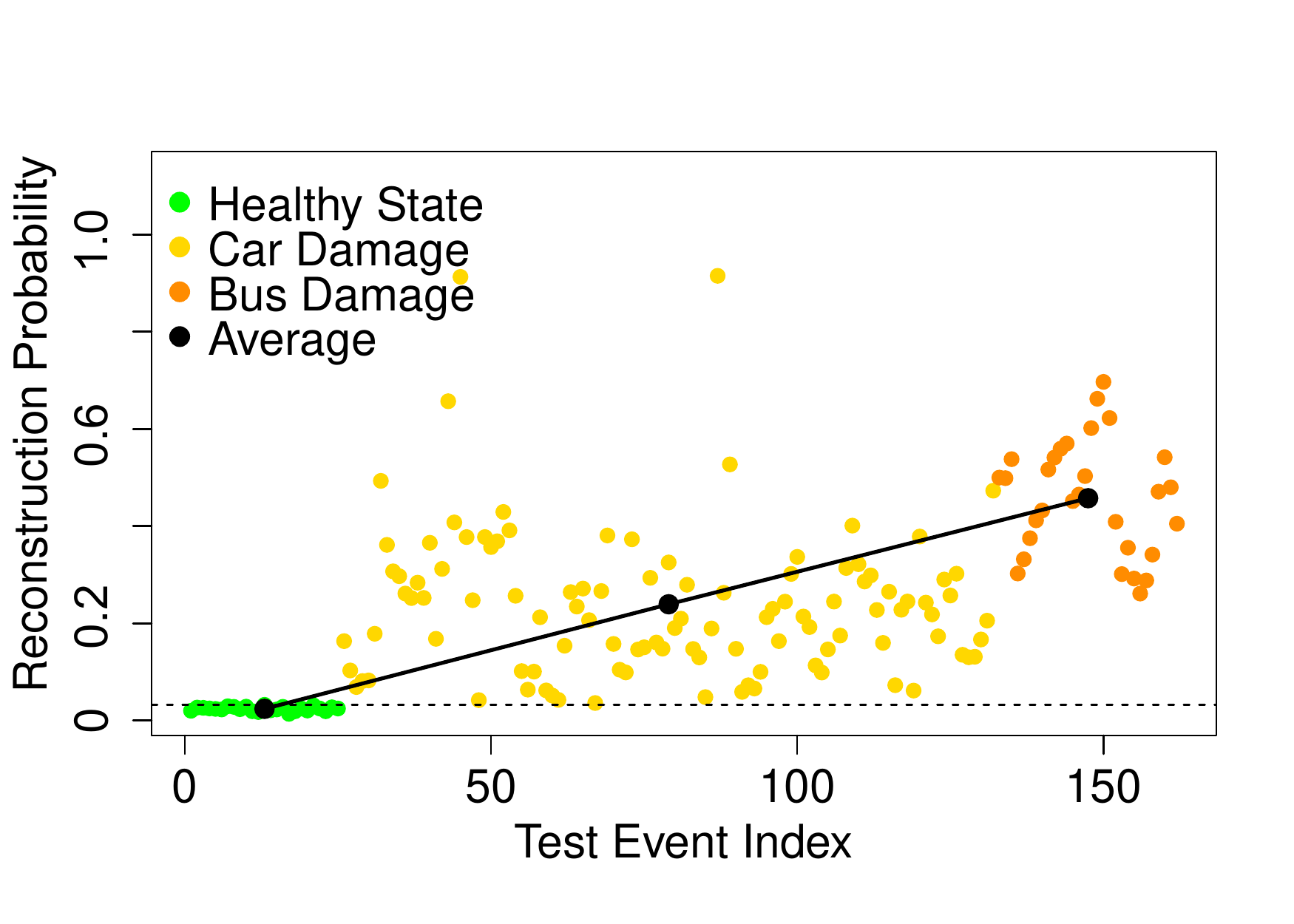}
	\caption{Damage estimation  using reconstruction probability values obtained by MAE applied  on the cable-stayed bridge dataset.}
	\label{fig:ae_wsu}
\end{figure}

We randomly selected eighty percent of the healthy events (100 samples) from each sensor to form  training multi-way of $ X \in \Re^{24 \times 600 \times 100}$ (i.e. \textit{training} set). The 137 examples related to the two  damage cases were added to the remaining 20\% of the healthy data to form a \textit{testing} set, which was later used for the model evaluation.  Our probabilistic anomaly detection algorithm was able to successfully detect all the healthy and damage events in the \textit{testing} data set, and achieved an F-Score of 100\%. Moreover, this model was able to assess the progress of  damage severity  in the structure based on the obtained  reconstruction probabilities. To illustrate that, we plotted the reconstruction probability  values for all test samples which were shown in Figure \ref{fig:ae_wsu}. The horizontal axis indicates the index of the test samples and the vertical axis indicates the magnitude of the reconstruction probability values. A value above the  horizontal dashed line  (as shown in Figure \ref{fig:ae_wsu}) indicates a sample classified as healthy, whereas a value below that line  indicates an event classified as damage. This line represents an anomaly threshold value which  was used to identify whether  a new  event  is belong to the healthy  or damage state.

The first 25 healthy events denoted by green dot were all correctly classified as healthy samples with a reconstruction probability below the anomaly threshold value of 3\% (97o\% of confidence interval). All the damage samples denoted by yellow and orange dot refer to the "Car-Damage" and "Bus-Damage", respectively,  generate  high reconstruction  probability values  above the anomaly threshold,  thus identified as damage.  We further calculated the mean of all the reconstruction probability  values for each state  to illustrate how the MVA model was also able to asses the severity of the identified damage.  Figure \ref{fig:ae_wsu} shows a solid black line which was drawn to connect the mean values. It can be clearly observed that the MVA  model was able  to  separate the two damage cases  ("Car-Damage" and "Bus-Damage") where the reconstruction  probability  values were further increased for the samples related to the more severe damage cases related to  "Bus-Damage".

\begin{figure}
	\centering
	\includegraphics[scale=0.5]{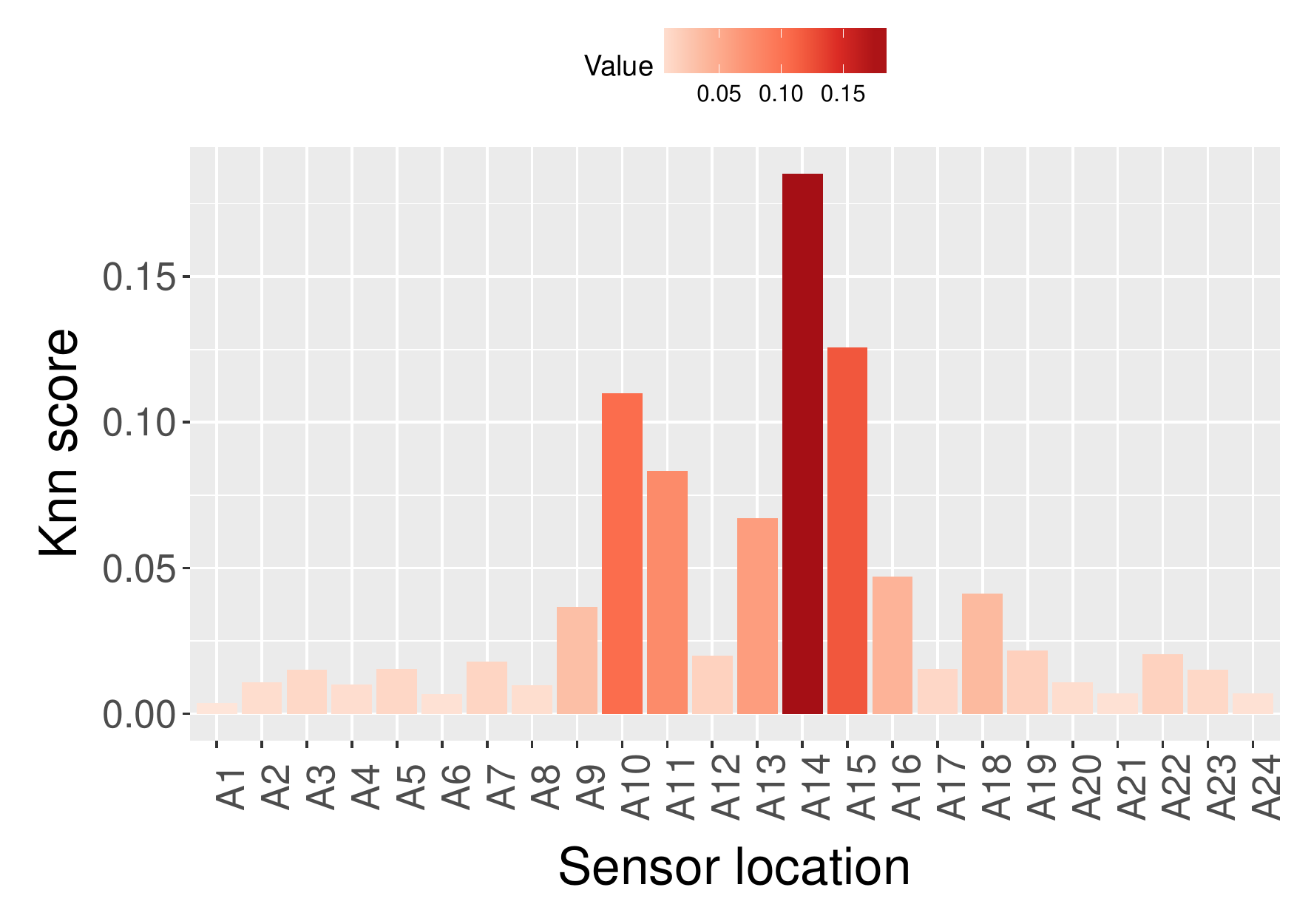}
	\caption{Location anomaly score in the localization layer applied on the cable-stayed bridge dataset.}
	\label{fig:local_wsu}
\end{figure}

The last step in MVA model was  to localize the position of the detected damage by analyzing the identity matrix $S_{new}$ where each row captures meaningful information for each sensor location. We calculated the average distance from each row in matrix $S$  to $k$-nearest neighbouring to $S_{new}$. The resultant $k$-nn score for each sensor is presented in Figure \ref{fig:local_wsu} which clearly shows the capability of MVA for   damage localization. As expected,   sensors A10 and A14   related to the  "Car-Damage" and "Bus-Damage", respectively, behaved  significantly  different from all the other sensors apart from the  position of the emulated damage.

\begin{figure}
	\centering
	\includegraphics[scale=0.4]{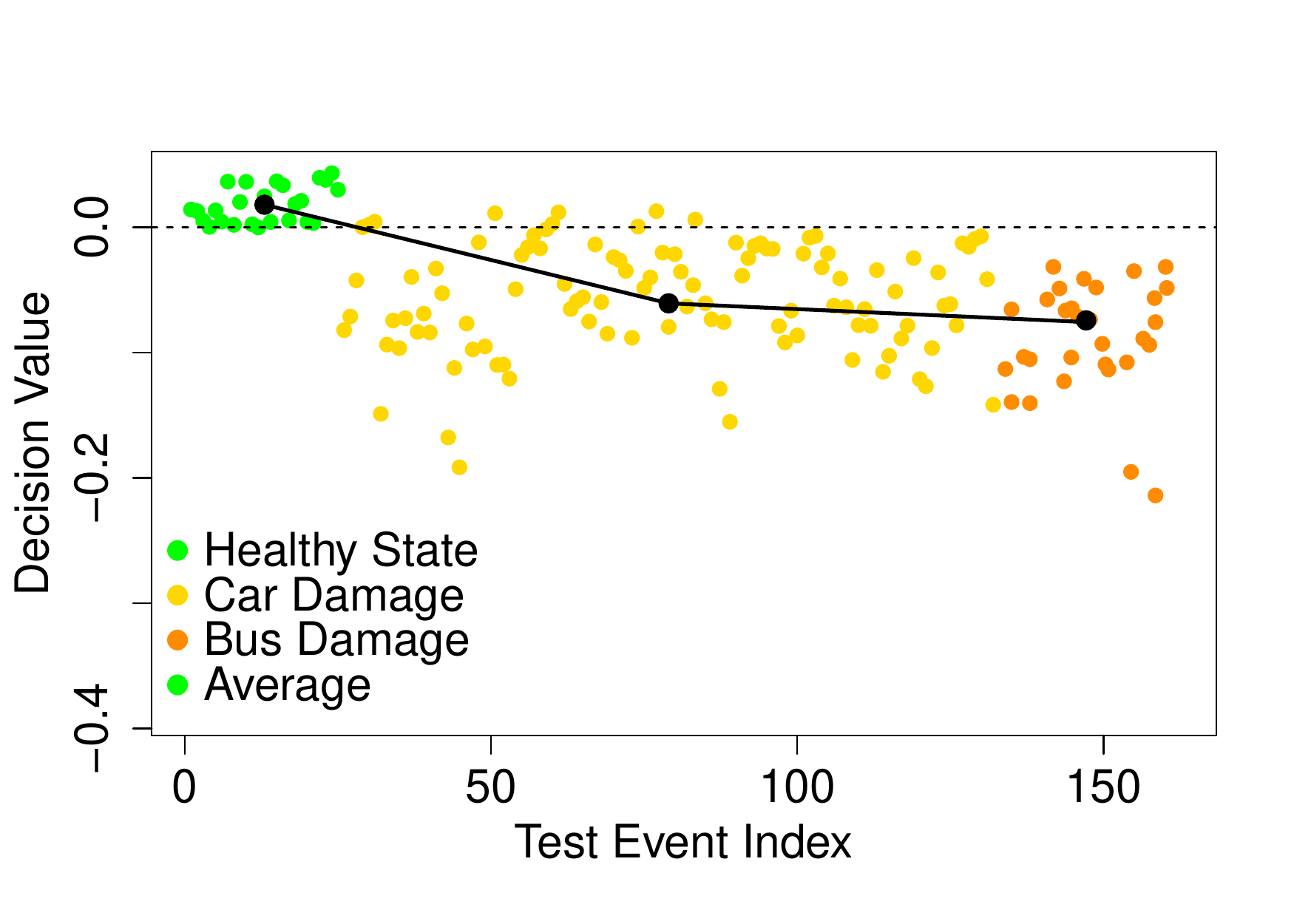}
	\caption{Damage identification results using OCSVM on the cable-stayed bridge dataset.}
	\label{fig:ocsvm_wpe}
\end{figure}

The next experiment was to compare our obtained results with the  state-of-the-art method   OCSVM.  The same \textit{training}  data set as above was used  to  construct a OCSVM model, and the same  \textit{testing} data set was used to evaluate the classification performance of OCSVM. The F-score accuracies of OCSVM was recorded at 95\%. However, the OCSVM decision values were not able to clearly assess the progress of the  damage severity  in the structure as illustrated in Figure \ref{fig:ocsvm_wpe}. Moreover, OCSVM is lacking the capability to implement a method for damage localization since  only one  single anomaly score for each event  is generated by OCSVM model using  the inputs from  sensors $\{A_i\}_{i=1}^{24}$. 

\subsubsection{Building Data}

Our second experiments were conducted using the acceleration  data  acquired from 24 sensors  instrumented on the three-story building  as described in Section \ref{s:data_b}. Similar to the previous experiments,  we  normalized the accelerometer data to have zero mean and unity variance. Then we applied FFT method to   represent the data in  frequency domain. For each two adjacent accelerometers at a location, we used   the difference between their signals as  variables and only the top 150Hz were selected as input features to our MVA model.  The resultant three-way  data  has a structure of 12 locations $\times$ 768 features $\times$ 240 events.

We randomly selected 80\% of the healthy events (120 samples) from the 12 locations as a training multi-way data $ X \in \Re^{12 \times 768 \times 120}$ (i.e.\textit{training} set). The remaining 20\% of the healthy data and the data obtained from the two damage cases were used for testing (i.e.\textit{testing} set). 

\begin{figure}
	\centering
	\includegraphics[scale=0.4]{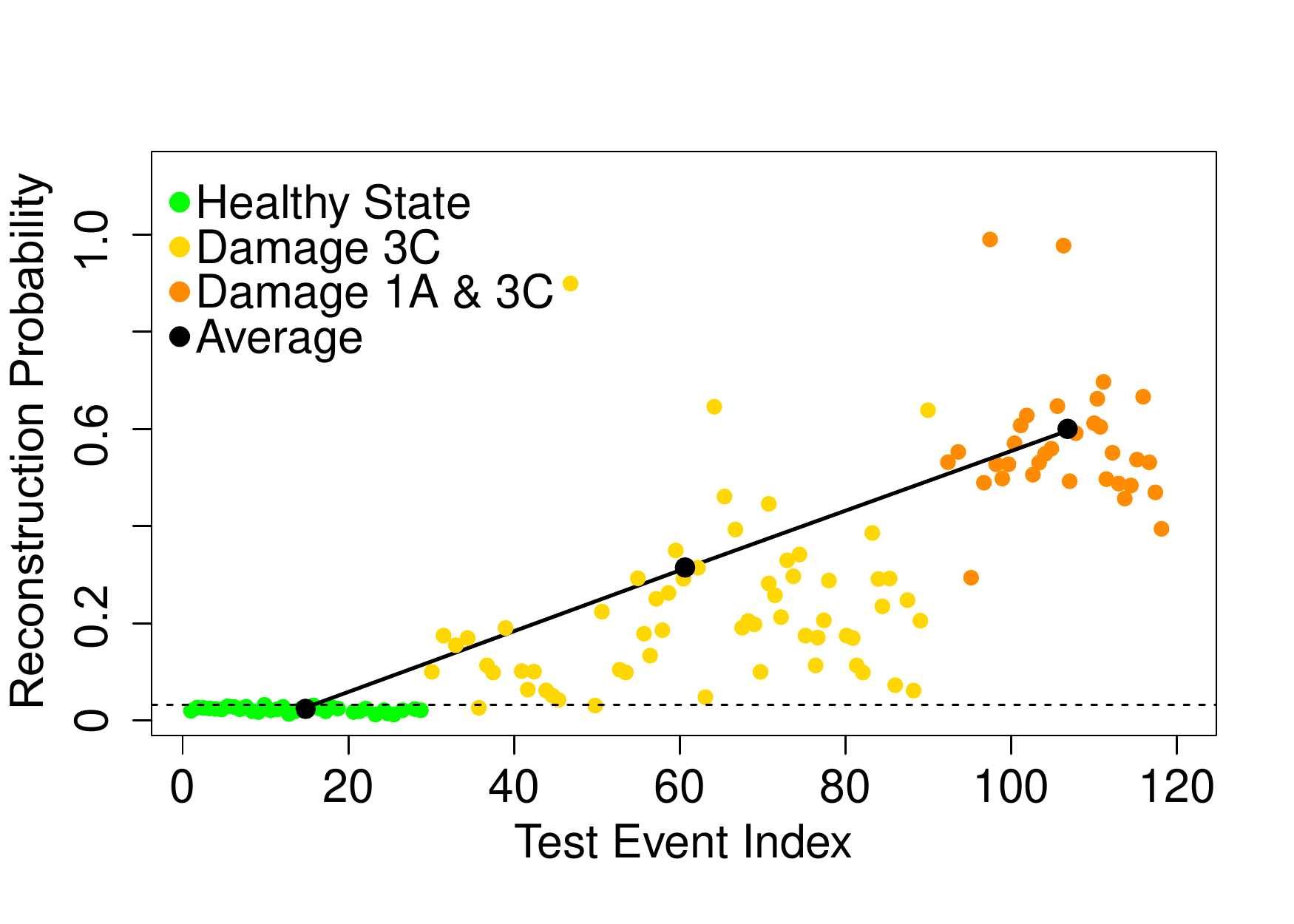}
	\caption{Damage estimation  using reconstruction probability values obtained by MAE applied on the Building dataset.}
	\label{fig:ae_build}
\end{figure}

Our constructed MVA model achieved an F-score of 97\%. The false alarm rate was equal to zero where all the healthy samples are correctly detected in the \textit{testing} data set.  Figure \ref{fig:ae_build}  shows the plot of the reconstruction  probability values  generated by   MVA. It can be clearly observed from Figure \ref{fig:ae_build} that the more severe damage test data related to  locations 1A and 3C  were more deviated from the training data with high reconstruction probability values.

\begin{figure}
	\centering
	\includegraphics[scale=0.4]{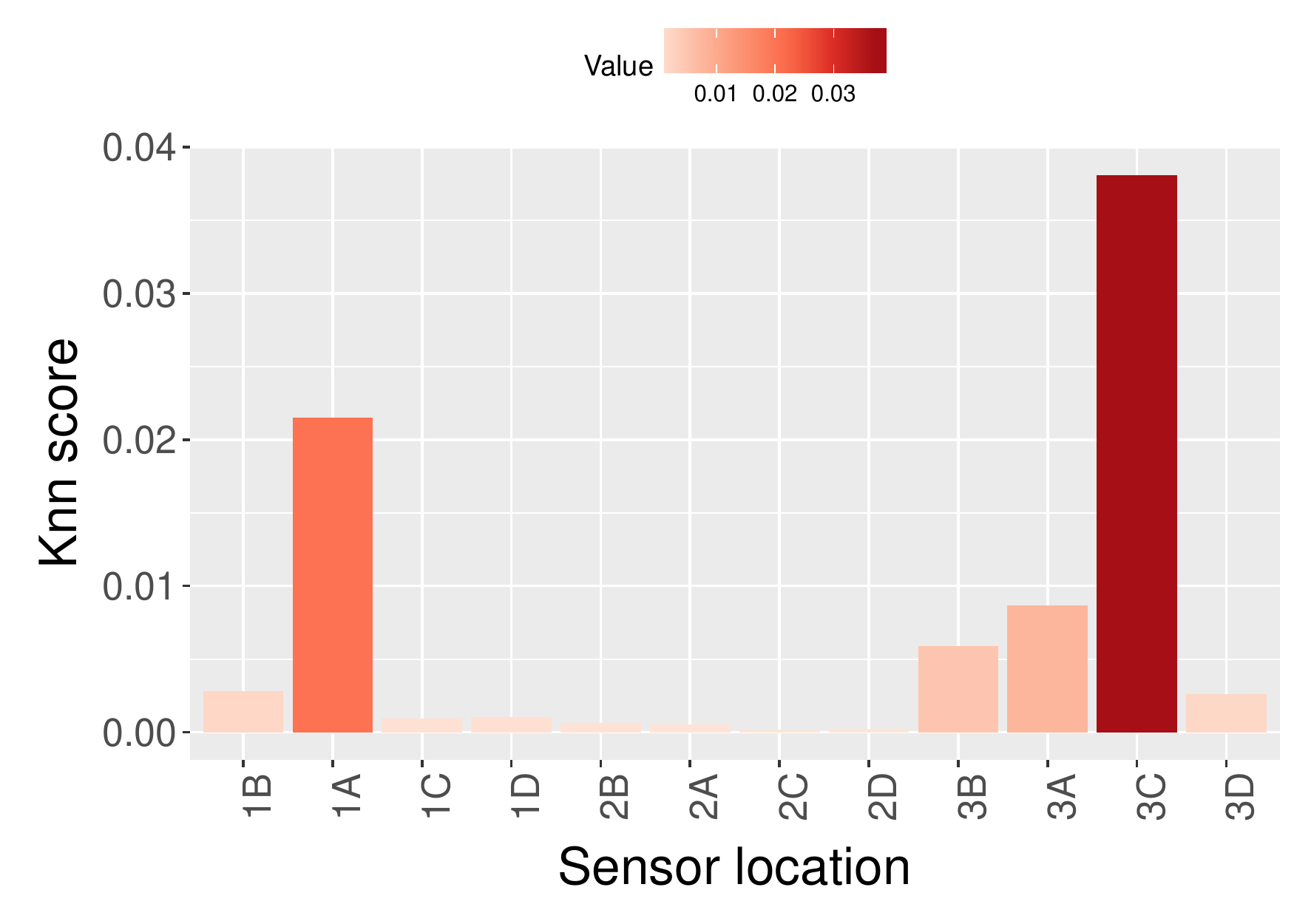}
	\caption{Location anomaly score in the localization layer on the Building dataset}
	\label{fig:build_local}
\end{figure}

Similar to the last case study, we further propagated the reconstruction probability values  obtained by  the output layer into  the localization layer to construct $S_{new}$ matrix. Then we computed the $k$-nn score for each sensor based on the  average distance between  each row of  matrix $S$  to $S_{new}$.  Figure \ref{fig:build_local}  shows the resultant $k$-nn score for each sensor.  It clearly shows that MVA method correctly captures damage locations. As expected,   sensors 1A and 3C  produced very high  $k$-nn score due the introduced damage at these two locations.  The $k$-nn score of 3C was higher than 1A because that damage was  introduced in both locations 1A and 3C at the same time. 

\begin{figure}
	\centering
	\includegraphics[scale=0.4]{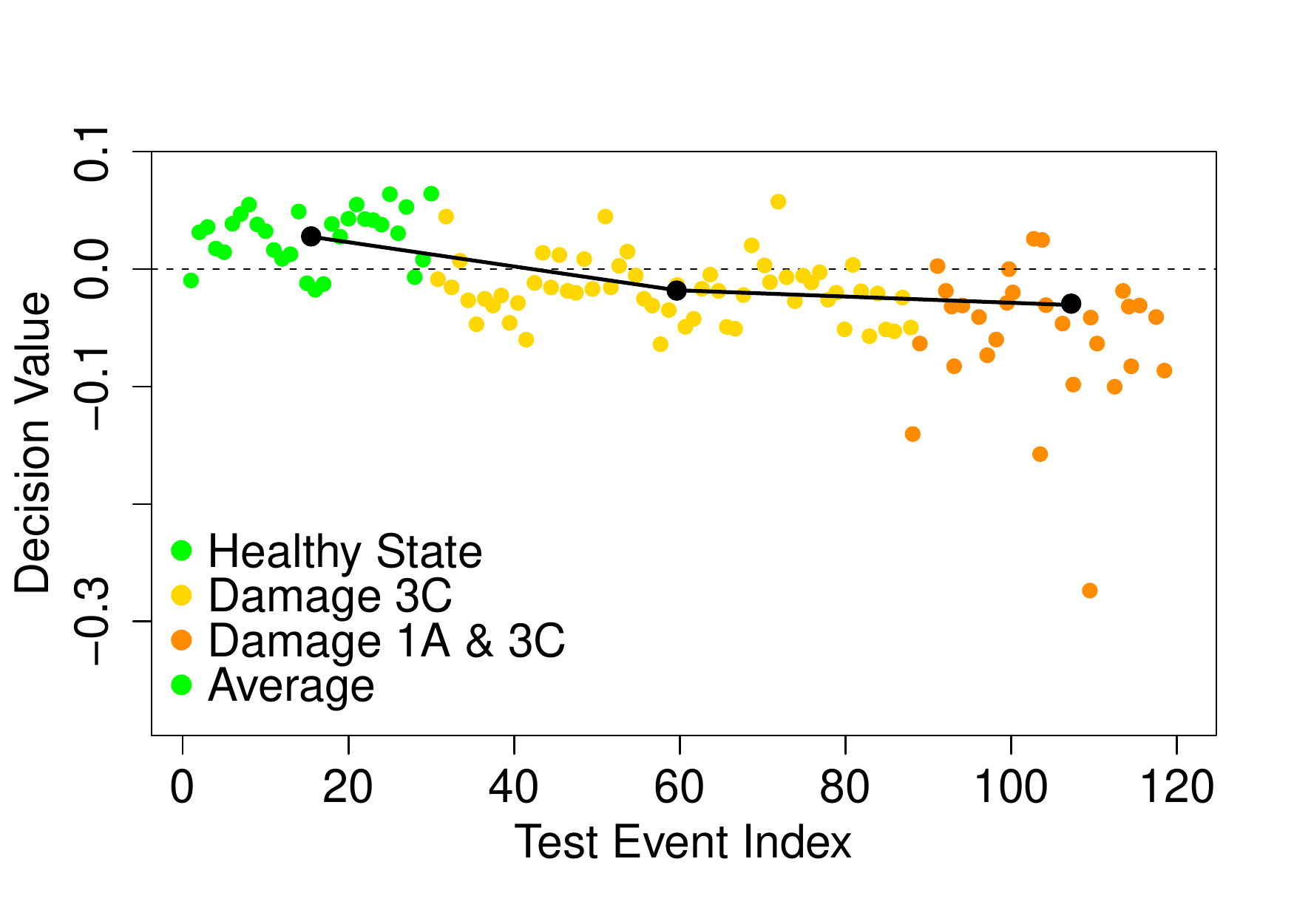}
	\caption{Damage estimation  using decision values generated by OCSVM applied on the Building dataset.}
	\label{fig:ocsvmbuild}
\end{figure}

The last  experiment was to compare our obtained results with OCSVM. The F-score accuracy of OCSVM was recorded at 86\% with no clear separation between  the different levels of damage as illustrated in Figure \ref{fig:ocsvmbuild}. Again, OCSVM doesn't have the  capability to implement a method for damage localization since  only one  single anomaly score for each event  is generated by OCSVM model using  input data  from  sensors $\{A_i\}_{i=1}^{24}$.

\section{Conclusion}
\label{s:conclusion}

Multiway data analysis has gained a lot of interest in many fields where standard two way analysis   don't have the capabilities  to learn underlying structure of the multi-way data. We  proposed  a multi-objective variational autoencoder method for damage detection, localization and severity assessment in  multi-way structural data   based on the reconstruction probability of the autoencoder deep neural network. The proposed method performs data fusion  by taking input features from a networked sensors attached to a structure. Stochastic  gradient descent  algorithm is then used to   learn reconstructions that are close to its original input slice followed by constructing a sensor identity matrix which used for damage localization. For each new incoming data slice we calculate its  anomaly score   based on reconstruction probability and we use the obtained reconstruction probability values  for damage assessment. The sensor identity matrix is finally utilized to locate the identified damage. 

We evaluated our method on multi-way datasets in the area of structural health monitoring for damage detection purposes. The data was collected from our deployed data acquisition system on a cable-stayed bridge in Western Sydney and  from a laboratory based  building structure obtained from Los Alamos National Laboratory (LANL). Experimental results showed that our approach succeeded at detecting  the damage events with an average F-score of 0.95\% and higher for all datasets. Moreover, Our  model demonstrated the capability  to work very well in localizing damage and estimating  different levels of damage severity  in an  unsupervised aspect. Compared to the state-of-the-art approaches, our proposed method shows better performance in terms of damage	detection and localization.

\begin{acks}	
The authors also would like to thank the Western Sydney University and University of New South Wales for facilitating the field tests and data collection process.
\end{acks}

\bibliographystyle{ACM-Reference-Format}
\bibliography{ae_ref}

\end{document}